# How can one sample images with sampling rates close to the theoretical minimum?

L. P. Yaroslavsky,
Dept. of Physical Electronics, School of Electrical Engineering,
Tel Aviv University, Tel Aviv 69978, Israel, yaro@eng.tau.ac.il

**Abstract**

A problem is addressed of minimization of the number of measurements needed for digital image acquisition and reconstruction with a given accuracy. A sampling theory based method of image sampling and reconstruction is suggested that allows to draw near the minimal rate of image sampling defined by the sampling theory. Presented and discussed are also results of experimental verification of the method and its possible applicability extensions.

## 1. Introduction

Sampling is the primary operation in acquiring digital images. Contemporary digital display devices and image processing software assume by default that sampling over regular square sampling grids is used for discrete representation of images. As it is well known, digital images acquired in this way are, as a rule, highly compressible. Hence images are compressed for storage and transmission and then, for displaying or processing, are reconstructed to the standard sampled representation.

The phenomenon of ubiquitous compressibility of images raises very natural questions: what are fundamental reasons of the compressibility of images sampled in a regular way and is it possible just directly measure the minimal amount of data that won't end up being thrown away? These questions were apparently first posed by the inventors of the Compressed Sensing approach as a solution to this problem [ 1], [ 2], [ 3]. This approach suggests methods of image reconstruction from lesser number of measurements than the required number of pixels by means of finding an image "sparse" approximation, i.e. an image with minimal number of non-zero spectral coefficients in the domain of a chosen "sparsifying" transform.

Compressed sensing approach is based on the well-known property of image "sparsifying" transforms, such as discrete cosine transform, wavelets, and other similar transforms, to compact most of image energy in a small amount of transform coefficients. It was proven in the theory of Compressed Sensing that if an image of $N$ samples is known to have, in domain of a certain transform, only $K$ non-zero transform coefficients out of $N$, the image can be precisely reconstructed from $M > K$ measurements by means of minimization of **L0** norm in the transform domain, provided the following inequality holds ([ 4] - [ 6]).

$$M/K > -2\log[(M/K)(K/N)] \qquad (1)$$

The number $K$ of signal non-zero transform coefficients is the theoretical minimum of the number of measurements required for signal reconstruction. The ratio $M/K$ of the number of required measurements $M$ to the number $K$ of signal non-zero transform coefficients represent sampling redundancy with respect to the theoretical minimum. One can obtain from inequality (1) that for the range of image spectrum sparsity $(K/N)$ of natural images from $10^{-1}$ to $2 \times 10^{-3}$, the required redundancy $M/K$ of the number of required measurements $M$ with respect to the theoretical minimum $K$ reaches 3 to 8 times ([ 7]). This means that compressed sensing is still far from reaching the theoretical minimum of signal sampling rate.

The reason of such sampling inefficiency of compressed sensing methods lies in full uncertainty regarding positions of signal non-zero coefficients in transform domain, which is assumed in the compressed sensing approach. This uncertainty can be partly resolved in many practical situations, because it is also well-known that, as a rule, properly designed "sparsifying" transforms compact most of signal energy into transform coefficients that correspond to lower image spatial frequencies. It was shown in [ 7] on an example, that, using this property of image transforms, one can reduce the sampling redundancy very substantially and reach sampling rates, which are sufficiently close to the theoretical minimum. An additional advantage of this approach is that it allows direct specification of resolution of reconstructed images.

In this paper we elaborate this idea in details and show that an efficient practical solution of the problem of minimization of the number of samples sufficient for image reconstruction with a given accuracy can be found by computational imaging means.  In Sect. 2 we remind basics of sampling theory and practical methods of image sampling in order to reveal reasons for sampling redundancy of conventional image sampling methods. In Sect 3, we use a discrete imaging model and the discrete sampling theorem to derive nearly non-redundant image sampling method that enables reaching sampling rates sufficiently close to the minimum determined by the sampling theory. In Sect. 4 results of extensive experimental verification of the method are provided. In Sect. 5 some practical issues of usage of the proposed method are addressed. In Sect 6 perspectives of using of the proposed approach for solving other under-determined inverse problems in digital imaging are briefly discussed. Conclusion summarizes the results.

## 2. Why conventional image sampling methods result in image oversampling?

Signal sampling is based on the idea of signal band limited approximation. Consider basics of signal and image sampling. For 1D signals, sampling is quite simple. Here is the signal sampling and reconstruction protocol:
- for a signal $s(x)$ to be sampled, define admissible Mean Square Error (MSE) $\sigma^2$ of its approximation;
- determine an interval $[-B, B]$ of the signal Fourier spectrum that contains $(E-\sigma^2)/E$ -th fraction of signal energy $E$ ;

- pass the signal through the ideal low-pass filter **LPF** with frequency bandwidth $[-B, B]$ to obtain a band-limited approximation $\tilde{s}(x)$ of signal $s(x)$:

$$\tilde{s}(x) = \int_{-\infty}^{\infty} s(\xi) LPF(x - \xi) d\xi \qquad (2)$$

and sample this signal with sampling interval $\Delta x = 1/2B$ to obtain signal samples $\{\tilde{s}(k\Delta x)\}$:

$$\tilde{s}(k\Delta x) = \frac{1}{\Delta x} \int_{-\infty}^{\infty} s(\xi) LPF(k\Delta x - \xi) d\xi, \quad k = ...., -2, -1, 0, 1, 2, ..... \qquad (3)$$

- For signal reconstruction, pass impulse signal $\left[ \sum_{k=-\infty}^{\infty} \tilde{s}(k\Delta x) \delta(\xi - k\Delta x) \right]$ through the ideal low-pass filter with frequency bandwidth $[-B, B]$ to produce the band-limited approximation $\tilde{s}(x)$ of the signal $s(x)$:

$$\tilde{s}(x) = \sum_{k=-\infty}^{\infty} \tilde{s}(k\Delta x) LPF(x - k\Delta x) \qquad (4)$$

$LPF(.)$ in above equations is point spread function of the low-pass filter.

Signal sampling rate $2B$ is the minimal rate that secures image reconstruction with the given reconstruction error variance, i.e. minimum $2BX$ samples is required for a signal of length $X$. Practical implementation of this optimal 1D sampling is limited only by technical problems of implementation of low pass filters that approximate the ideal low pass filter.

One can use a similar protocol for images as 2D signals as well:
- for a given image $s(x, y)$, choose sampling intervals $(\Delta x, \Delta y)$ (or sampling rate, in "dots per inch") over a regular rectangular sampling grid;
- obtain signal samples:

$$\tilde{s}(k\Delta x, l\Delta y) = \frac{1}{\Delta x \Delta y} \int_{-\infty}^{\infty} s(\zeta, \eta) LPF(k\Delta x - \xi, l\Delta y - \eta) d\xi \quad k = ...., -2, -1, 0, 1, 2, ....., l = ...., -2, -1, 0, 1, 2, ....., \qquad (5)$$

using the 2D ideal low pass filter with bandwidth $(-1/2\Delta x, 1/2\Delta x; -1/2\Delta y, 1/2\Delta y)$

- For image reconstruction, pass 2D impulse signal $\left[ \sum_{k=-\infty}^{\infty} \tilde{s}(k\Delta x, l\Delta y) \delta(\xi - k\Delta x) \delta(\eta - l\Delta y) \right]$ through the ideal low-pass filter with frequency bandwidth $(-1/2\Delta x, 1/2\Delta x; -1/2\Delta y, 1/2\Delta y)$ to produce a band-limited approximation $\tilde{s}(x, y)$ of the image $s(x, y)$:

$$\tilde{s}(x, y) = \sum_{k=-\infty}^{\infty} \sum_{l=-\infty}^{\infty} \tilde{s}(k\Delta x, l\Delta y) LPF(x - k\Delta x, y - l\Delta y) \qquad (6)$$

with approximation MSE defined by the image energy outside the band-limiting rectangle $(-1/2\Delta x, 1/2\Delta x; -1/2\Delta y, 1/2\Delta y)$.

This, in fact, is how commonly used image sampling and display devices work. The role of sampling and, correspondingly, image reconstruction low-pass filters is, as a rule, played by image sensor and image display devices apertures. Sampling intervals $(\Delta x, \Delta y)$ are chosen usually on the base of knowledge of resolving power of imaging optics (in "lines per mm") or of visual assessment of images, assuming, say, that for reproducing the sharpest edge in the image one needs at least 2 samples. The image sampled representation standard assumes that sampling is carried out over square or rectangular sampling grids.

According to the sampling theory, the minimal number $N_{min}$ of samples sufficient for reconstruction of images from their samples with MSE $\sigma^2$ is equal to the product $S_{xy} B_\Omega$ of the image area $S_{xy}$ and the area $B_\Omega$ of the zone $\Omega$ in the Fourier domain that contains, by virtue of the Parseval's identity, $(E - \sigma^2)/E$-th fraction of the image signal energy $E$. We will call these spectral zones that contain a chosen fraction of image signal energy Energy Compaction (EC-) zones. If image is sampled over a square sampling grid with sampling intervals $(\Delta x, \Delta y = \Delta x)$, the square base band $(-1/2\Delta x, 1/2\Delta x; -1/2\Delta x, 1/2\Delta x)$ must encompass the image spectrum EC-zone. Therefore the number of image samples on a square sampling grid $S_{xy}/\Delta x^2$ will inevitably always exceed this minimal number of samples: $S_{xy}/\Delta x^2 \geq N_{min} = S_{xy} B_\Omega$.

Figure 1 illustrates this assertion on examples of 5 test images and their Fourier spectra. Fourier spectra were estimated using Discrete Fourier Transform and applying to images, before spectral analysis, a circular apodization mask in order to smoothly bring a sampled image down to zero at the edges of the sampled region and in this way to avoid as much as possible estimation errors due to boundary effects. EC-zones of image spectra that contain 99.5% of image energy are highlighted in the figures. MSE of image reconstruction from frequency components within the highlighted zones is of the order of image JPEG compression error. Relative areas of image EC-zones with respect to the square base-band, i.e. spectra sparsity, range between 0.195-0.265, which means that images are roughly 4-5 times oversampled.

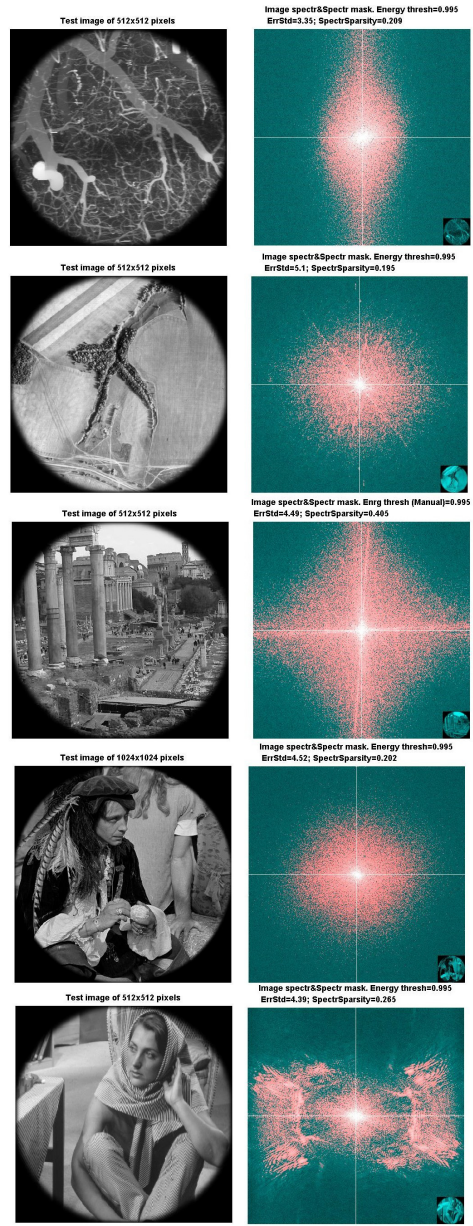

Figure 1. A set of sampled images (left column) and their corresponding Fourier spectra centered at their DC component (right column). Highlighted are spectra EC-zones that contain 99.5% of image energy.

In what follows we show that one can, using means of computational imaging, sample images with much lesser redundancy and reach sampling rates that are quite close to the theoretical minimum.

## 3. Nearly non-redundant image sampling method

Consider a discrete imaging model. Let $N$ be the number of image samples required for image representation over a standard regular square sampling grid. Assume that available are $K < N$ image samples and it is required to reconstruct the entire image of $N$ samples with minimal MSE. Choose an image transform $\Phi_N$ and use available $K$ image samples to compute $K$ transform coefficients. Then a "bounded spectrum" approximation to the entire image can be reconstructed by the inverse transform of the set of found $K$ transform coefficients with the rest of $N - K$ coefficients set to zero. Reconstruction MSE will be equal to the energy of $N - K$ coefficients set to zero. For a given transform, this error can be minimized if $K$ largest transform coefficients are used for the reconstruction. In order to further minimize image reconstruction error, one should choose a transform with a better capability of energy compaction into the small number of transform coefficients. The said is the meaning of the Discrete Sampling Theorem ([ 8]).

Note that in the limit, when $N \to \infty$, the discrete model converts to the continuous one. In particular, if Discrete Fourier transform is chosen as the image transform, it converts to the integral Fourier transform and the Discrete Sampling Theorem converts to the classic sampling theorem.

The Discrete sampling theorem, theoretically, suggests a possible solution of the problem of minimization of the number of image samples sufficient for image reconstruction with a given accuracy. To make this option practical, four issues must be resolved:
- Choosing a transform.
- Specifying the ways image samples have to be taken.
- Specifying image spectra EC-zone in transform domain, i.e. of the subsets of transform coefficients to be used for image reconstruction.
- Method of image reconstruction from the obtained sampled representation.

Consider possible resolutions of these issues.

<u>Choosing a transform</u>. The choice of the transform is governed by the transform energy compaction capability. An additional requirement is the availability of a fast transform algorithm. From this point of view, Discrete Fourier Transform (DFT), Discrete Cosine Transform (DCT) and wavelet transforms are among primary candidates.

<u>Specifying positioning of image samples</u>. Positioning of samples should permit computation, from image samples, of the group of transform coefficients chosen for image reconstruction. Some image transforms, such as wavelets, impose certain limitation on positions of image samples. For DFT and DCT, positions of image samples can be arbitrary ([ 8]). An additional advantage of using DFT and DCT as image sparsifying transforms is that they are discrete representations of the integral Fourier transform and, as such, they ideally concord with characterization of imaging systems in terms of their Modulation Transfer Functions.

<u>Specification of image EC-zones</u>. Specification of the subset of transform coefficients to be used for image reconstruction, i.e. of the EC-zone in transform domain, can be made on the basis of the known capability of image transforms, such as DCT, to compact most of the image signal energy into few transform coefficients that form in the transform domain more or less compact groups in the area of lower indices around the DC component. Practical experience, including that obtained in course of developing of zonal quantization tables for image compression standards, such as JPEG, shows that although these groups do not have sharp borders, they are quite well concentrated. This means that the groups can be, with a reasonably good accuracy in terms of preservation of the group total energy, circumscribed by one of some compact standard shapes that encompass area of image low spatial frequencies and can be specified by few parameters, such as area, aspect ratio, angular orientation. Figure 2 presents a set of possible standard shapes suited for DCT as the sparsifying transform: rectangle, pie-sector, ellipse and super ellipse. In principle, each particular standard shape can be associated with a certain class of images, such as micrographs, aerial photographs, space photos, in-door and out-door scenes, etc.

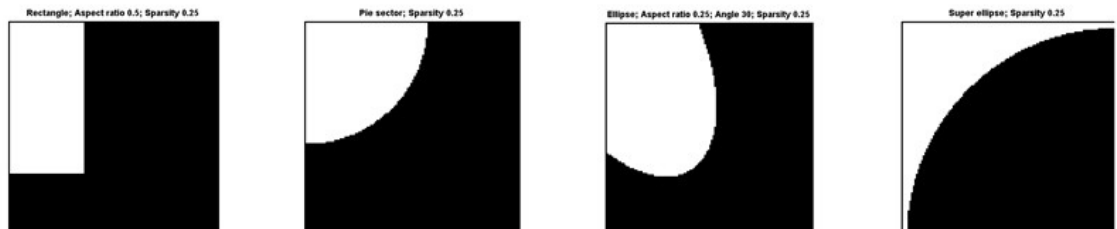

Figure 2. Examples of possible simple standard shapes for image DCT spectra. From left to right: rectangle, pie-sector, ellipse, super ellipse. Spectrum DC component is in the upper left corners of the shapes.

The author's experimental experience shows that no fine tuning of shape parameters is required for specifying shape parameters. This property of sparse DCT spectra is illustrated in Figure 3 on a sparse DCT spectrum of a test image shown in Figure 3, a). One can easily notice in the image a certain prevalence of horizontal edges. This prevalence causes anisotropy of image sparse spectrum seen in boxes b) – d), where shown are marked as white dots non-zero DCT coefficients that reconstruct this image with root mean square (RMS) error 3.85 gray levels of 255 levels (36.4 dB), the same as the reconstruction RMS error of this image after its standard JPG compression by the Matlab tools. Additionally in boxes b) – d) shown are borders of oval and rectangular shapes that have the same area (0.275 of the total area) and different aspect ratios (0.3, 0.45 and 0.35, correspondingly). When used as spectrum bounding shapes, they all reconstruct the test image with practically the same RMS reconstruction errors (3.8, 3.8 and 4.1 of image gray levels correspondingly). As one can see in Figures 4 a) and b) reconstructed images obtained for two cases of spectrum bounding, by oval, as in Figure 3, b), and by rectangle, as in Figure 3 d), are visually indistinguishable one from another, though patterns of the reconstruction errors (Figures 4, c) and d)) look, naturally, a bit different (for display purposes reconstruction errors are shown 8 times contrasted).

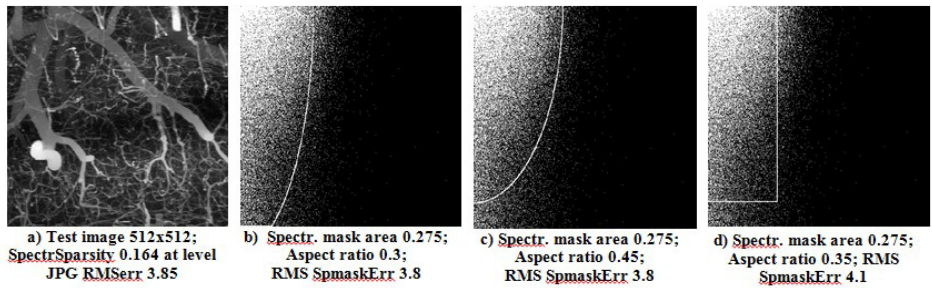

Figure 3. Test image 512x5122, spectrum sparsity 0.164 at level JPEG RMS error 3.85 (a), its nonzero spectral components for the reconstruction error RMS 3.85 of image gray levels (white dots) and borders (white lines) of approximative oval and rectangular shapes with different shape parameters ( b) –d)).

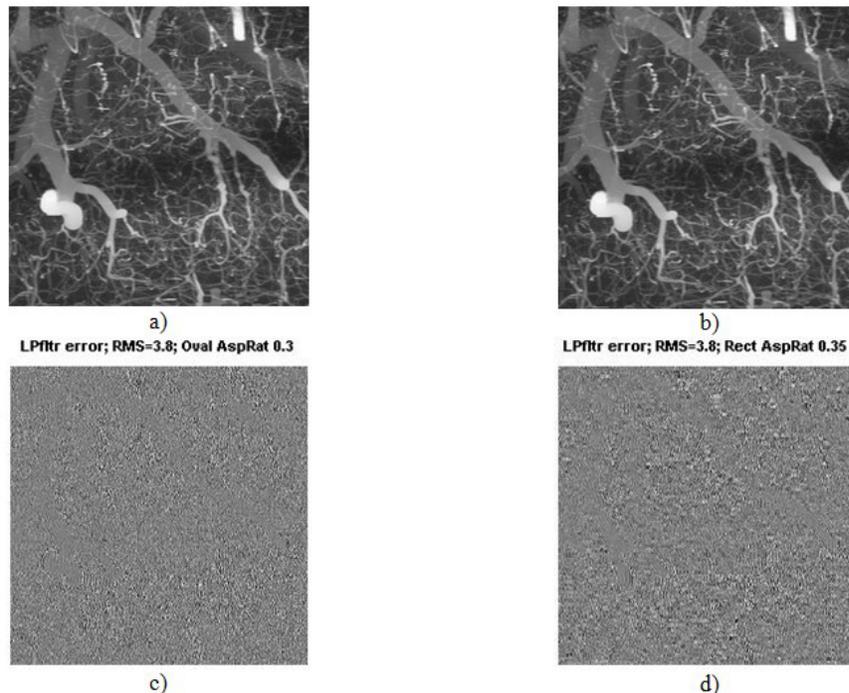

Figure 4. Images (a, b) reconstructed from test image of Figure 3, a) after its band limitations by oval (Figure 3, b) and by rectangular (Figure 3, d) spectral masks and patterns of corresponding reconstruction errors (c, d, displayed 8 times contrasted)

As one can see from the maps of image sparse spectrum non-zero components in Figure 3, shapes that are chosen to encompass the EC-zone of spectral coefficients to be used for image reconstruction, do not include all non-zero spectral coefficients of the sparse spectrum and, from the another side, include some of its zero components. Zero spectral components that happen to be inside of the chosen spectral shape have, by definition, lower energy than non-zero components that happen to be outside the chosen spectral shape. Therefore, given the energy of all spectral components encompassed by the chosen spectral shape, the number of these internal zero components must exceed the number of non-zero components not encompassed by the shape. This means that the area of the shape that defines the number of samples to be taken by the necessity exceeds to a certain degree the number of sparse spectrum non-zero coefficients, which, theoretically, is the minimal number of samples required. For instance redundancies of EC-zones of the image in Figure 3 are 0.275/0.164=1.67. Experimental experience reported in the next section shows that redundancies of standard EC-zones for real images are, as a rule, of the same order of magnitude. This EC-zone redundancy is the price one should pay for the uncertainty regarding exact indices of transform coefficients to be used for reconstruction of the particular image with a given accuracy.

*Method of image reconstruction*. For image reconstruction, one can consider two options:

- Direct inversion of the $K \times N$ inverse transform matrix that links $K$ available samples and $K$ transform non-zero coefficients specified by the selected spectral EC-zone with $N - K$ coefficients set to zero. Once $K$ non-zero transform coefficients are found and the rest $N - K$ transform coefficients are set to zero, the inverse transform is applied to the found spectrum to reconstruct all $N$ samples. Generally, matrix inversion is a very hard computational task and no fast matrix inversion algorithms are known. In our specific case, a pruned fast transform matrix should be inverted. There exist pruned versions of fast transforms for computing subsets of transform coefficients of signals with all samples except several ones equal to zero [9]), which is inverse to what is required in the given case. The question, whether these pruned algorithms can be inverted for computing a subset of transform non-zero coefficients from a subset of signal samples is open.

- An iterative Papoulis-Gerchberg type algorithm. The algorithm at each iteration consists of two steps:

  (i) The iterated reconstructed image is subjected to the direct transform and then the obtained spectrum is multiplied by the chosen bounding EC-zone mask for obtaining the iterated transform spectrum.

  (ii) The iterated transform spectrum is inverse transformed and samples of the obtained image at positions, where they were actually taken at sampling, are replaced by the corresponding available samples, which produces the next iterated reconstructed image.

As a zero order approximation, from which reconstruction iterations start, an image interpolated in one or another way from the available samples can be taken (the interpolation algorithm used in verification experiments is detailed in next section).

The above reasoning suggests the following protocol of image sampling assuming DCT as the image sparsifying transform:

- Choose the highest required image spatial resolution $SpR$ (in "dots per inch") in the same way as it is being done in the conventional image sampling.
- On the basis of evaluation of the image to be sampled, choose one of the standard spectral bounding shapes for bounding EC-zone of DCT spectrum and its shape parameters, such as aspect ratio for rectangle and super ellipse, aspect ratio and orientation angle for ellipse, etc.
- For the chosen shape, evaluate the fraction $Fr$ of the area the shape occupies in the square, which circumscribes it; this fraction times $SpR \times SpR$ determines spatial density $SpD$ of samples to be taken (in "dots per square inch"): $SpD = Fr \times SpR \times SpR$. The number of samples $M$ to be taken can be then found as a product of $SpD$ and the image area $ImgSzX \times ImgSzY$ : $M = SpD \times ImgSzX \times ImgSzY$.
- Choose whatever sampling grid appropriate for the available image sensor and sample the image; if no other option is available, use sensor's aperture as a pre-sampling low-pass filter.

For image reconstruction:

- Choose the number of image samples $N > M$ over a dense uniform square sampling grid intended for image reconstruction; $N$ should be sufficiently large to secure accurate positioning physical sampling positions at nodes of the uniform sampling grid.
- Apply to the sampled image one of the described reconstruction procedures using for specification of the image spectrum EC-zone the chosen spectrum bounding shape. In this way an image with spectrum in the chosen transform bounded by the chosen EC-zone, or a bounded spectrum (BS-) image, will be obtained, which approximates the sampled image with the given accuracy.

As one can see, the described sampling protocol does not essentially differ from the conventional standard 2D sampling protocol. The only difference is that in the suggested method arbitrary sampling grids can be used and evaluation of image expected spectrum shape for choosing spectrum EC-zone is required in addition to the evaluation of image resolution, which anyway is required by the standard sampling protocol. Not much different is image reconstruction from sampled data as well. In the suggested method, low pass filtering at image reconstruction is carried out numerically by means of bounding image spectrum in the chosen transform by the chosen spectral shape. Thanks to this, the method reaches the minimal sampling rate defined by the area of the chosen spectral shape, although the latter is somehow larger than the area of the image sparse spectrum, which it approximates and which defines the absolute minimum of the sampling rate. In this sense, the suggested image sampling method can be called Nearly Non-redundant (NNR)-sampling method.

## 4. Experimental verification of the method

The suggested image NNR-sampling method has been experimentally verified on a considerable data base of test images from the USC-SIPI Image Database ([ 10]). In the experiments, the above-described iterative Gerchberg-Papoulis type algorithm was used and three types of sampling grids were tested: (i) quasi-uniform sampling grid, in which $M$ image samples are uniformly distributed, with appropriate rounding off their positions to the nearest nodes of a dense square sampling grid of $N$ samples; (ii) uniform sampling grid with jitter, in which horizontal and vertical positions of each of $N$ samples are randomly chosen, independently in each of two image coordinates, within the primary uniform sampling intervals; and (iii) random sampling grid, in which positions of samples are uniformly and totally randomly distributed over the dense sampling grid of $N$ samples. An image transform that compacts the image spectrum, the Discrete Cosine Transform, was used. As an admissible root mean square (RMS) error of approximation of test images by images with sparse DCT spectra, RMS errors of image compression by the standard JPEG compression in Matlab implementation are taken. As a zero-order approximation, from which the iterative reconstruction starts, each not available image sample was interpolated from three nearest to it available samples taken with weights inversely proportional to their distance from the interpolated sample.

Figure 5 and Figure 6 illustrate results of experiments with five images of the tested set. Shown in Figure 5 are: (i) test image, (ii) reconstructed image, (iii) sampled test image, (iv) border of the chosen shape of image EC-zone (super ellipse) and positions of image DCT spectrum non-zero coefficients (white dots) that contain spectral coefficients, which reconstruct image with RMS error equal to that of image JPEG compression; (iv) plots of RMS of 90% lowest reconstruction errors and RMS of the total reconstruction error vs iteration number. Separate count of 90% of smallest reconstruction errors was motivated by the observation that iterative reconstruction converges not uniformly over the image area: most of the errors decay with iterations much more rapidly than few isolated large errors. RMS of reconstruction errors are given in units of image gray levels in the range 0-255.

In Figure 6 shown are, for the sake of saving space, only reconstructed images (left column), maps of non-zero coefficients of sparse approximation to the corresponding test images and borders of their chosen EC-zones (middle column) and plots of RMS of reconstruction errors versus the number of iterations (right column).

For all shown images, sampling over uniform sampling grids with jitter was used, for which reconstruction errors decay most rapidly. For the same number of iteration, RMS of reconstruction errors for random sampling grid is about 1.5-2 times and for quasi-uniform sampling grid 2-2.5 times larger than those for "uniform with jitter" sampling grid. For "quasi-uniform" sampling grids, stagnation of the iteration process was observed, which can apparently be attributed to the presence of regular patterns of thickening and rarefication of sampling positions due to rounding off their coordinates to positions of nodes of the regular uniform sampling grid.

Numerical results for reconstruction accuracy, redundancy of the chosen EC-zones (the ratio of the fraction of area they occupy in the spectral domain to the spectrum sparsity), sampling redundancy (the ratio of the sampling rate to the minimal sampling rate defined by the area of the EC-zone) and overall sampling redundancy (the ratio of the sampling rate to the spectrum sparsity) obtained for above shown five test images are presented in Table 1.

Table 1. Summary of experiments

| Test image | Reconstruction error RMS (PSNR) | Redundancy | | |
|---|---|---|---|---|
| | | EC-zone | Sampling | Overall |
| Rome 512 | 1.94 (42.4 dB) | 1.6 | 1 | 1.6 |
| Barbara 512 | 4.33 (35.4 dB) | 1.6 | 1 | 1.6 |
| | 0.69 (51.4 dB) | 1.6 | 1.15 | 1.84 |
| Pirat 1024 | 1.15 (46.9 dB) | 1.61 | 1 | 1.61 |
| Aerial photo 512 | 1.21 (46.5 dB) | 1.54 | 1 | 1.54 |
| Blood vessels 512 | 1.94 (42.3 dB) | 1.61 | 1 | 1.61 |

As one can see in Figure 5 and Figure 6, plots of RMS reconstruction errors vs the number of iterations show that RMS of reconstruction errors decays at first couple of hundreds iterations quite rapidly but after it reaches the value of 2-3 quantization intervals, the error decay slows down. It was found in the experiments that the error decaying can be substantially accelerated if the number of samples is taken with a certain redundancy, i.e.

10-20% larger than the minimal number equal to the area of the chosen EC-zone (see results for test image "Barbara 512" in Table 1). This, of course, correspondingly increases the overall sampling redundancy.

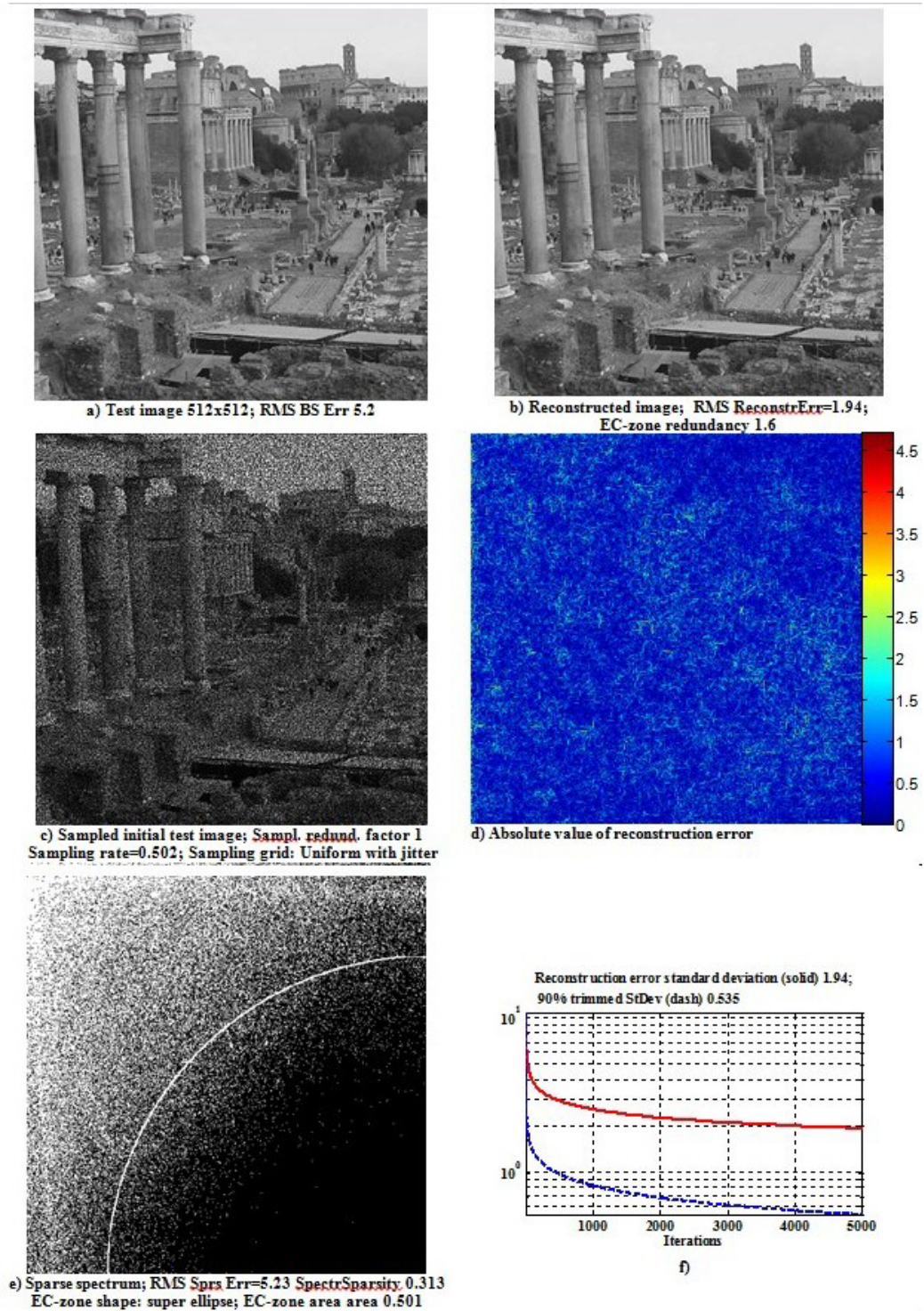

Figure 5. Results of experiments on NP-sampling and BS-reconstruction of test image "Rome512": a) – test image; b)- reconstructed BS-image; c) – sampled test image; d) – test image sparse spectrum (white dots) and the border of the chosen EC-zone (white solid line); e) – color coded (Matlab color map "jet") absolute value of reconstruction error (difference between images (a) and (b)). f) – plot of RMS of reconstruction error vs the number of iterations.

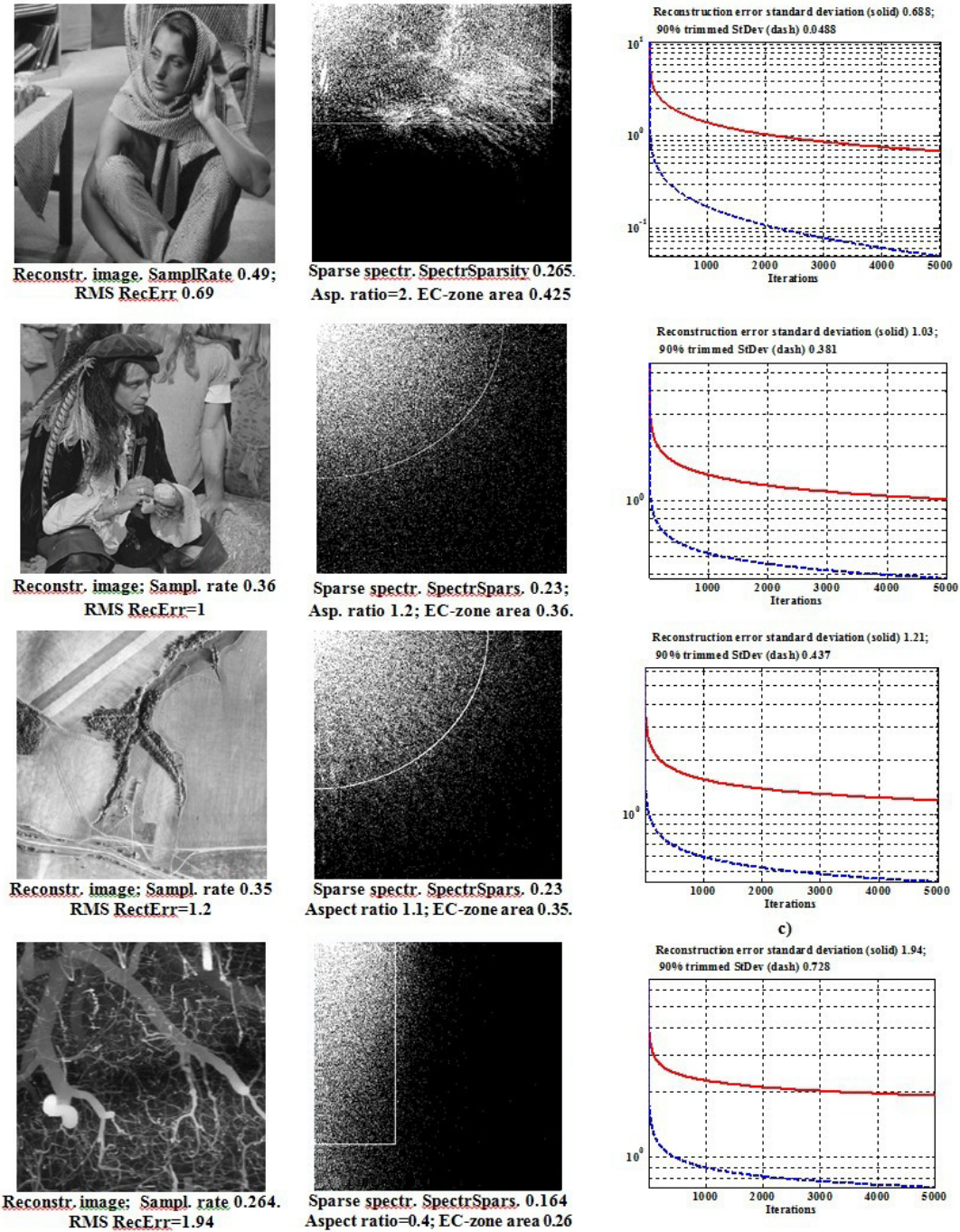

Figure 6. Results of experiments on sampling and reconstruction of test images, from top to bottom, "Barbara 512" (512x512 pixels), "Pirat 1024" (1024x1024 pixels), "Aerial photo" (512x512 pixels) and "Micrograph Blood vessels" (512x512 pixels). From left to right: reconstructed images, image sparse spectra (white dots) and borders of the corresponding chosen EC-zones (white solid line) and plots of RMS of reconstruction errors vs the number of iterations

To summarize, the experiments confirm that images sampled with sampling rate equal to the minimal rate for chosen EC-zones of images can be reconstructed with a sufficiently good accuracy. The redundancy in the number of required samples associated with redundancy of standard shapes of image spectra EC-zones is of the order 1.5 and never exceeded 2 in experiments with other images. These figures are estimates of the overall sampling redundancy of the suggested NNR-sampling method.

## 5. Some practical issues

In this section, three issues of practical application of the suggested image NNR-sampling method are addressed: (i) how robust is the method to the presence of noise in sensor data; (ii) practical considerations on the shape of EC-zone for bounding image spectra for image reconstruction and (iii) computational complexity of the method.

An important practical issue of applicability of the suggested image NNR-sampling and BS-reconstruction method in practical imagers is whether it is robust to the presence of noise in the image sensor. From the method description in section 3 one can see that the method, just as the conventional sampling and reconstruction, is linear, i.e. it satisfies the superposition principle. No parameter of sampling and reconstruction algorithms depends on signal values and, in particular, on whether noise is present in the signal or not. Therefore sampling and reconstruction of an image that contains additive noise will result in a reconstructed image that also contains additive noise with power spectrum modified by the frequency response of the sampling device and bounded by the shape of the spectrum EC-zone used for image reconstruction. If the sensor noise is white with variance $\sigma^2$ and the frequency response of the sampling device is flat in the base band, noise in the reconstructed image will be not white and will have variance $\kappa\sigma^2$, where $\kappa < 1$ is the relative area of the reconstruction EC-zone. Because in reality sampling devices perform certain low-pass filtering of images, the degree of reduction of noise variance in the reconstructed images will be correspondingly larger.

In order to illustrate the said, an experiment on sampling and reconstruction of an image with and without additive noise was conducted. The results are presented in Figure 7, from which one can see that the presence of noise in sampled data has no influence on the work of the reconstruction algorithm.

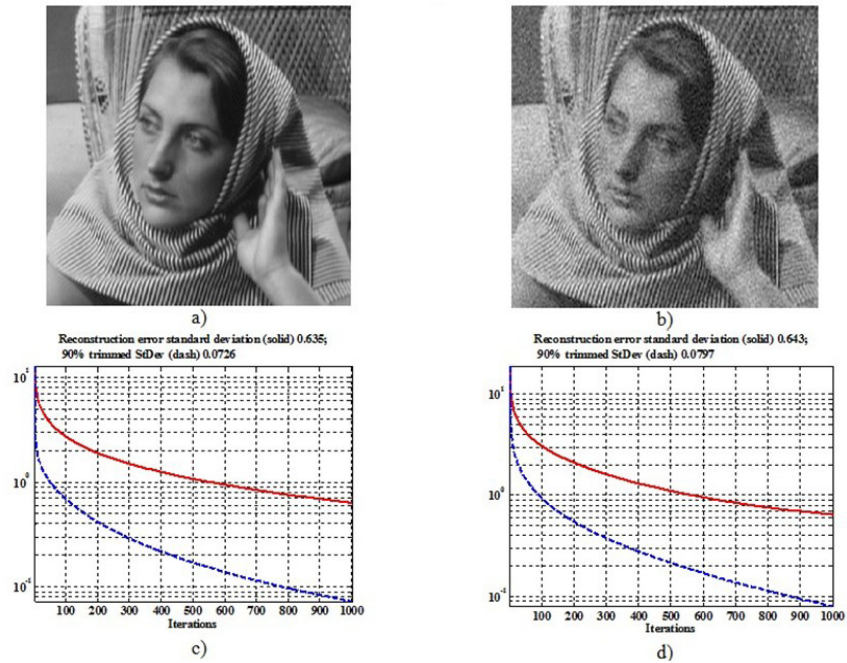

Figure 7. Images reconstructed from a sampled noiseless test image (a) and from same sampled image with added white Gaussian noise with standard deviation 20 gray levels (b) and the corresponding graphs (c, d) of RMS of reconstruction errors versus the number of iterations.

Another practical issue is selecting the shape of EC-zone for bounding image spectra for image reconstruction. As it was mentioned in Sect. 3 no fine tuning is required for specifying shape parameters. Therefore it is suggested that several standard shapes should be found for different classes of images such as landscape, portrait, micrographs, aerial and space photographs of different kind and alike. This can be based, for instance, on a machine learning algorithm trained on various image data bases. For sampling images in a particular application, the user should only specify the image class. Note that specifying an image class is the standard option for setting parameters of modern digital cameras.

The applicability of the proposed method depends also on its computational complexity. The computational complexity of the reconstruction algorithm per iteration is determined by the complexity $O(2N\log N)$ of floating point operations needed for direct and inverse fast transforms plus $O(N)$ replacement operations needed for sample wise modifications of data in image domain ($M$ operations) and in its transform domain ($N-M$ operations). The order of magnitude of time required for one iteration can be estimated from these data: elapsed time for Matlab direct or inverse DCT of an array of $N = 512\times 512$ numbers implemented on a PC "Lenovo-201" with processor Intel i7 and operating system Windows-7 is 52 msec.

## 6. Other possible applications of image NNR-sampling and BS-reconstruction

The above discussed problem of reconstruction of images of $N$ samples from $M < N$ sampled data can be considered as a special case of under-determined inverse problems. One can expect that the found solution of this problem, the bounded spectrum (BS-) reconstruction of images, may find application for solving other under-determined inverse problems as well. We illustrate this possibility in three applications: (i) in-painting of occlusions in images, (ii) image reconstruction from sampled spectra and (iii) image reconstruction from modulus of its Fourier spectrum.

Image in-painting. In-painting of image occlusions can be carried out with exactly the same algorithm as the above described iterative algorithm of image BS-reconstruction from sparse samples, in this case those samples that are not occluded. An illustrative example is presented in Figure 8.

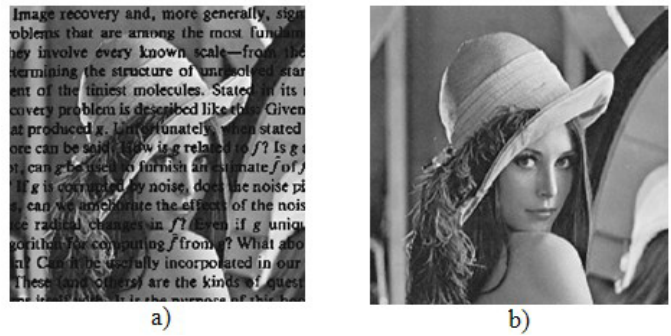

Figure 8. An example of using image BS-reconstruction for image in-painting: a) a test image with occlusions; b) reconstructed image; RMS of reconstruction error is 2.15 gray levels (PSNR 41.5 dB)

Sampling in spectral domain. There exist some imaging devices (e.g. some healthcare scanners), where sampling is done in a transform domain. The proposed NNR sampling and BS-reconstruction method can be used in such devices, if it is known, as it frequently happens in tomography, that object image is surrounded by some empty space. Figure 9 demonstrates this option on an example of image reconstruction from its sampled Fourier spectrum. In this example, Fourier spectrum of a test image bounded by a circular binary image mask was randomly sampled with sampling rate equal to the ratio of the image bounding circle area to the area of the entire image frame. Additionally, spectrum was bounded by a circular binary spectral mask with radius equal to the highest horizontal and vertical spatial frequency of the baseband. This gives an additional $\pi/4 = 21.5\%$ saving in the number of spectrum samples.

For image reconstruction, an iterative algorithm was used. At each iteration, the iterated spectrum is Fourier transformed for obtaining an iterated reconstructed image and then the latter is multiplied by the bounding circular image mask and inverse Fourier transformed. Samples of the obtained spectrum in positions of original ones are replaced by them, and the spectrum is bounded it by the circular binary spectral mask to form the iterated spectrum for the next iteration. As a spectrum zero approximation for the iterative reconstruction, the sparsely sampled and bounded spectrum was used.

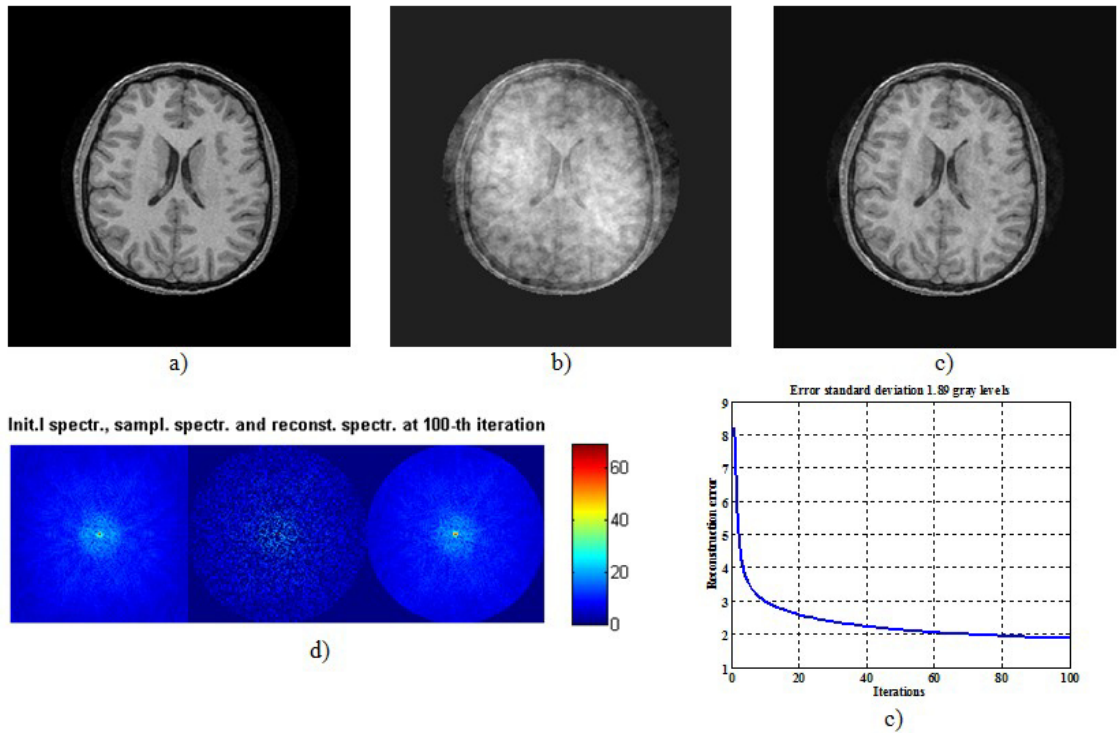

Figure 9. Image reconstruction from sampled spectrum: a) - test image bounded by a binary circular mask with radius equal to 0.35 of entire image size; b) - reconstructed image at first iteration, in which the circular bounding mask can be seen; c) - reconstructed image at 100-th iteration; reconstruction error standard deviation is 1.89 (PSNR=49 dB); d), from left to right: Fourier spectrum of the test image, its spectrum randomly sampled with sampling rate $\pi \times 0.35^2 \times \pi/4 = 0.3$ and reconstructed spectrum (for display purposes, absolute values of spectral samples are displayed raised to power 0.3); e) – plot of reconstruction error standard deviation vs the number of iterations.

<u>Image reconstruction from modulus of its Fourier spectrum</u>. In order to enable image reconstruction from modulus of its Fourier spectrum, using the suggested NNR-sampling and BS-reconstruction method, object should be imaged through a randomized binary (opaque-transparent) mask that produces occlusions in the object image. Fraction of the transparent area of the mask should be equal to or larger than the required by the NNR-sampling method sampling rate, i.e. the fraction of area occupied in the spectrum base band by the selected for this image EC-zone of the image DCT spectrum. Measured is module of Fourier spectrum of the object occluded by the mask. Reconstruction of the entire object image is conducted in two stages. At the first stage, object image with occlusions is reconstructed from modulus of its Fourier spectrum using an iterative algorithm of Gerchberg-Papoulis type. Iterations reconstruct the image with occlusions and estimates of phase component of its Fourier spectrum. At each iteration, a current spectrum phase component estimate is combined with the measured spectrum modulus to form a complete spectrum estimate, which is inverse Fourier transformed to obtain a current estimate of the reconstructed image with occlusions. Then this image is multiplied by the binary mask, which restores occlusion, and Fourier transformed. Phase component of obtained spectrum is used as the next estimate of image phase component, and iterations are repeated. As a zero order estimate of the image spectrum phase component, phase component of the Fourier spectrum of the binary mask can be used. At the second stage of image reconstruction, the reconstructed image with occlusions is used for BS-reconstruction of the entire image in the same way as in the above described image in-painting. An illustrative example is shown in Figure 10

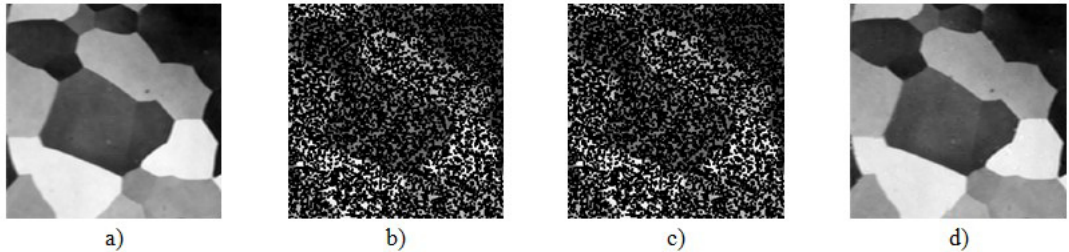

Figure 10. Image BS-reconstruction from modulus of its Fourier spectrum : a) a test image; b) the test image masked by randomly placed 3x3 pixel opaque squares; c) masked image reconstructed from modulus of its Fourier spectrum; RMS of reconstruction error is $10^{-3}$ (PSNR 107 dB). d) image BS-reconstructed from the reconstructed masked image (c); RMS reconstruction error is 1.79 (PSNR 43.1 dB).

## 7. Conclusion

The problem of minimization of the number of measurements required for image reconstruction with a given accuracy is addressed. It is shown that ubiquitous compressibility of digital images in their standard sampled representation over square sampling grids roots in disparity of the sampling grids and shapes of Fourier spectra of natural images. A "nearly non-redundant" (NNR-) method of image sampling and numerical reconstruction is proposed that enables to draw near to the minimal sampling rate defined by the sampling theory. The method assumes representing images in the domain of one of transforms with a sufficiently good energy compaction capability, approximating EC-zones of transform coefficients, which secure image reconstruction with a given accuracy, by one of the standard shapes and reconstructing images with spectrum bounded by the chosen shape (BS-reconstruction).

Presented results of experimental verification of the method using DCT as the image transform demonstrate workability of the method. They also show that practical sampling redundancy of the method with respect to the ideal sampling, which assumes exact knowledge of image spectrum EC-zone for each particular image, is of the order 1.5-2. With respect to the class of images with the chosen standard EC-zone, the method is non-redundant. In addition, it secures the chosen resolving power of reconstructed images defined by the chosen EC-zone and is robust to the presence of noise in sensor data.

Examples of application of the image NNR-sampling and BS-reconstruction method for image in-painting, image reconstruction from sampled Fourier spectrum and image reconstruction from modulus of its Fourier spectrum demonstrate its potential applicability for solving other underdetermined inverse problems.